\documentclass[letterpaper, 10 pt, conference]{ieeeconf}  

\IEEEoverridecommandlockouts                              
\usepackage[utf8]{inputenc}
\usepackage[T1]{fontenc}

\usepackage{xcolor}
\newcommand{\revised}[1]{\textcolor{black}{#1}}
\usepackage[utf8]{inputenc}   
\usepackage[T1]{fontenc}      

\usepackage{amsmath,amsfonts}

\usepackage{bm}
\usepackage{algpseudocode}
\usepackage{algorithm}

\usepackage[caption=false]{subfig}
\usepackage{tikz}
\usetikzlibrary{arrows.meta, automata, positioning}

\usepackage{array}
\usepackage{textcomp}
\usepackage{stfloats}
\usepackage{url}
\usepackage{verbatim}
\usepackage{graphicx}

\usepackage{booktabs}
\usepackage{tabularx}

\usepackage{siunitx}

\usepackage{cite}
\usepackage{flushend}  

\makeatletter
\let\NAT@parse\undefined
\makeatother

\usepackage[colorlinks,linkcolor=blue,anchorcolor=black,citecolor=blue,urlcolor=blue,hyperfootnotes=true]{hyperref}   
\usepackage[all]{hypcap} 

\title{\LARGE \bf
Six-DoF Hand-Based Teleoperation for Omnidirectional Aerial Robots
}

\author{Jinjie Li$^{1\dag*}$, Jiaxuan Li$^{2\dag}$, Kotaro Kaneko$^{1,3}$, \revised{Haokun Liu$^1$}, Liming Shu$^{2}$, and Moju Zhao$^{1}$
\thanks{$^\dag$Equal contributions. $^*$Corresponding author.}
\thanks{$^{1}$DRAGON Lab, Department of Mechanical Engineering, The University of Tokyo, Tokyo 113-8654, Japan.
        {\tt\small \{jinjie-li, \revised{haokun-liu}, chou\}@dragon.t.u-tokyo.ac.jp}}%
\thanks{$^{2}$Intelligent Equipment and Medical Device Laboratory, Department of Mechanical Engineering, Dalian University of Technology, Dalian 116024, China.
        {\tt\small \{1592890733@mail.,l.shu@\}dlut.edu.cn}}%
\thanks{$^{3}$HNL Lab, Department of Mechanical Engineering, The University of Tokyo, Tokyo 113-8654, Japan.
        {\tt\small kaneko@hnl.t.u-tokyo.ac.jp}}%
\thanks{This work was supported in part by JSPS KAKENHI under Grant 23H03472 and in part by CSC.}
}

\begin{document}
\bstctlcite{IEEEtran:BSTcontrol}  

\maketitle
\thispagestyle{empty}
\pagestyle{empty}

\begin{abstract}

Omnidirectional aerial robots offer full 6-DoF independent control over position and orientation, making them popular for aerial manipulation. Although advancements in robotic autonomy, \revised{human operation} remains essential in complex aerial environments. Existing teleoperation approaches for multirotors fail to fully leverage the additional DoFs provided by omnidirectional rotation. Additionally, the dexterity of human fingers should be exploited for more engaged interaction.
In this work, we propose an aerial teleoperation system that brings the \revised{rotational flexibility} of human hands into the unbounded aerial workspace. Our system includes two motion-tracking marker sets---one on the shoulder and one on the hand---along with a data glove to capture hand gestures. Using these inputs, we design four interaction modes for different tasks, including \textit{Spherical Mode} and \textit{Cartesian Mode} for long-range moving, \textit{Operation Mode} for precise manipulation, as well as \textit{Locking Mode} for temporary pauses, where the hand gestures are utilized for seamless mode switching.
We evaluate our system on a \revised{vertically mounted} valve-turning task in \revised{the} real world, demonstrating how each mode contributes to effective aerial manipulation. 
This interaction framework bridges human dexterity with aerial robotics, paving the way for enhanced aerial teleoperation in unstructured environments.

\end{abstract}

\section{Introduction}


We humans continually aspire for robots to perform more powerful and precise operations over larger areas. Although traditional robot manipulators are designed to deliver greater strength and sometimes higher precision, their working spaces remain limited by the reach of their arms—much like human arms. To overcome these constraints, aerial manipulation is gaining popularity due to its ability to operate freely in 3D space, particularly in high-altitude environments that are difficult for humans to access. This advancement reduces risks, time, and costs for human workers \cite{ollero_past_2022, zhao_forceful_2022, nishio_design_2024}.

\begin{figure}[t]
    \centerline{\includegraphics[trim=0 0 0 0,clip,width=3.4in]{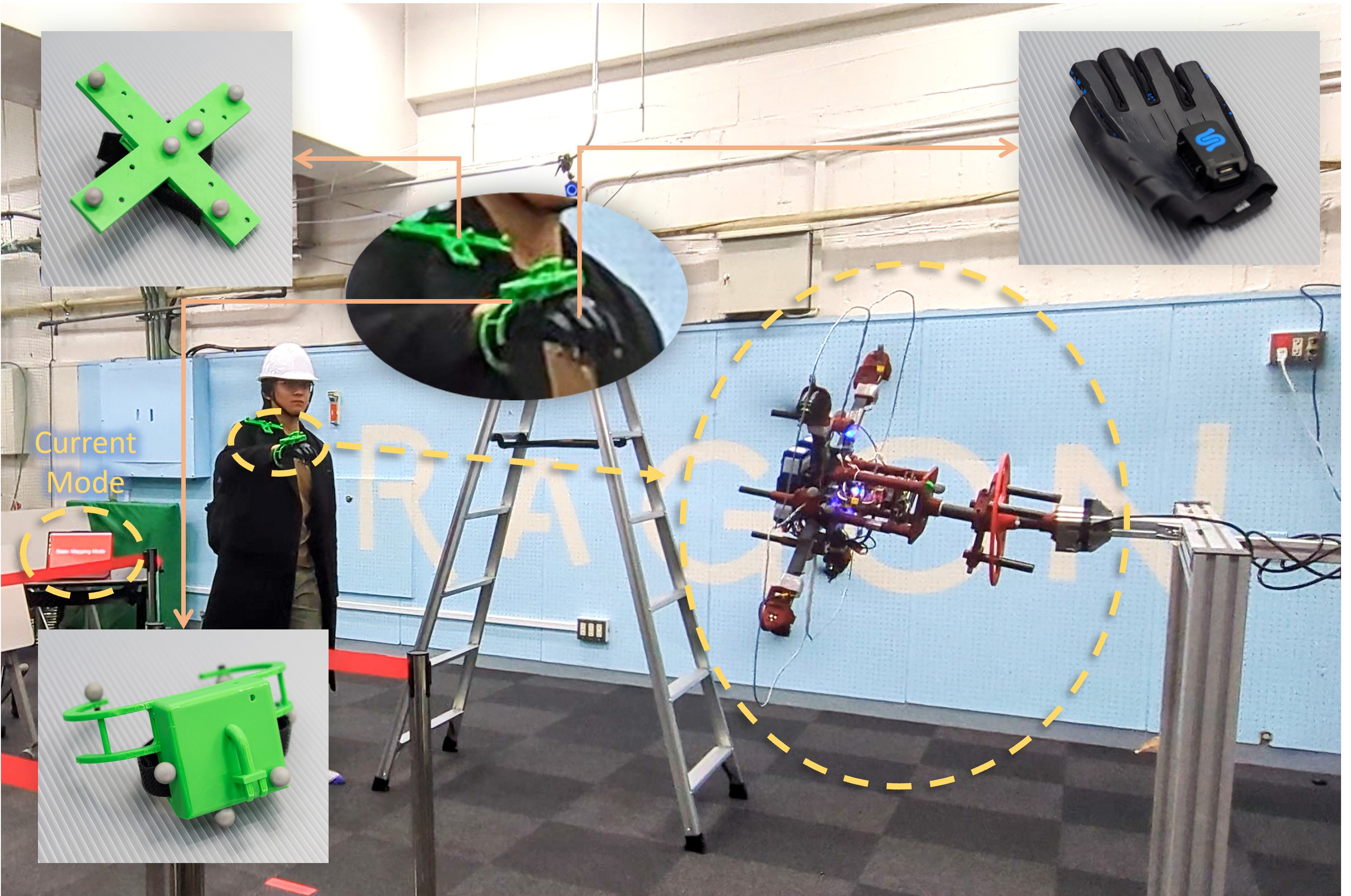}} 
    \vspace*{-2mm}
    \caption{\textbf{A human operator is controlling an omnidirectional aerial robot to turn a \revised{vertically mounted} valve.} The robot is in \textit{Operation Mode}, where its target pose is directly mapped from the operator's hand pose. \revised{A data glove is used to capture hand gestures for mode transitions.}
    \revised{The full video is released in
    \href{https://youtu.be/n0IQEnjPzrw?si=c5iDsgLRGipmvZTq}{https://youtu.be/n0IQEnjPzrw?si=c5iDsgLRGipmvZTq}}.}
    \label{fig:teleop_real}
\end{figure}

Although fully autonomous manipulation remains the ultimate goal for robotics researchers, the complexity of the aerial environment necessitates keeping a human operator in the loop \cite{darvish_teleoperation_2023}. Successful manipulation (such as flying into a factory and turning a valve) requires not only precise control and reliable structure, but also the ability to perceive the \revised{manipulated objects in the} environment, understand their physical properties, \revised{handle uncertainties,} and make sound decisions. 
Despite significant advancements in artificial intelligence, systems still struggle with spatial cognition and physical understanding. Therefore, teleoperation is attractive since it combines the physical part of a robot with the cognitive part of the human mind. Additionally, involving a human in the loop enhances safety by providing real-time oversight, which is critical for manipulation in disturbance-prone aerial environments \cite{mersha_bilateral_2014}. In this work, \revised{aiming to explore the sweet spot between human intelligence and machine capability,} we design a hand-based teleoperation system for omnidirectional aerial robots as shown in Fig.\;\ref{fig:teleop_real}.


\setlength{\textfloatsep}{8pt plus 1.0pt minus 2.0pt}

\revised{The core \revised{challenge} of aerial teleoperation is how to use the limited range of human motion to control the unrestricted movement of aerial robots.}
Researchers have designed various teleoperation methods for conventional multirotors \cite{tezza_state---art_2019, gioioso_flying_2014, kim_human-drone_2020, macchini_hand-worn_2020, tsykunov_dronestick_2021}. \revised{In most cases, operators use button-based input (e.g., keyboard) or joystick-based input—such as radio control (RC) systems or gamepads—to control drones.}
Most RCs and gamepads have 2 sticks, each with 2 degrees of freedom (DoFs), \revised{where the offset in each DoF typically corresponds to velocity.} \revised{This configuration results in} an operating \revised{space} of $\mathbb{R}^2 \times \mathbb{R}^2$, \revised{which is} sufficient to control 3D translation and yaw rotation for conventional multirotors. For aerial manipulation, researchers have developed various omnidirectional aerial robots that can generate forces in all directions and hover at any angle \cite{hamandi_design_2021, allenspach_design_2020, kamel_voliro_2018, brescianini_design_2016}, thereby extending the challenge to the $SE(3)$ space. Controlling these robots requires two additional inputs for roll and pitch, presenting challenges for traditional RC controllers. Furthermore, \revised{even if we add two more DoFs to the RC controller as in \cite{mellet_design_2025}}, the linear nature of a joystick cannot directly represent rotation. In other words, $\mathbb{R}^2 \times \mathbb{R}^2 \times \mathbb{R}^2 \cong \mathbb{R}^6$ is not equal to $SE(3) \cong \mathbb{R}^3 \times SO(3)$. Hence, using sticks to control orientation requires more cognitive load for coordinate transformation, making it more challenging than controlling translation \cite{dejong_improving_2004}.


To design an intuitive interaction method for omnidirectional aerial robots, researchers have tried human hands to directly indicate the target position and orientation. For example, in \cite{allenspach_towards_2022}, the operator holds the end of a robot arm to enable the bilateral teleoperation of an omnidirectional drone. Additionally, users in \cite{mellet_evaluation_2024} and \cite{lee_introspective_2024} hold a haptic device integrated with Virtual/Augmented/Mixed Reality interfaces to remotely control an aerial manipulator. However, although the robot is theoretically omnidirectional, the demonstrated tasks are limited to near-horizontal operations, such as pushing or insertion. This limitation may stem from the awkwardness of using a robot arm or the restricted operating range of a haptic \revised{device}. In contrast, we propose a system that uses the shoulder as the origin and maps the human hand's 6-DoF movements to an omnidirectional aerial robot, enabling \revised{challenging} tasks such as turning a vertically installed valve (Fig.\;\ref{fig:teleop_real}).


Another drawback of these hand-based methods \cite{gioioso_flying_2014, tsykunov_dronestick_2021, allenspach_towards_2022, mellet_evaluation_2024, lee_introspective_2024} is that they overlook a key feature of human hands: finger agility. \revised{The studies \cite{macchini_hand-worn_2020, yiu_skin-interfaced_2025} use hand gestures} to control the drone, but their results are limited to motion control rather than manipulation. Aerial manipulation involves more than just pushing or inserting; it also includes tasks like turning devices off, grasping objects, and executing complex maneuvers. Freeing the fingers for teleoperation broadens the range of possible operations in aerial manipulation. Moreover, the additional DoFs provided by the fingers allow for the design of more intuitive interaction methods, offering greater flexibility in human-robot interaction. As an initial trial \revised{to explore this concept}, we use finger gestures to switch among different control modes, enabling users to \revised{actively} select the most appropriate mode for each task. \revised{The hand gesture can be further utilized in the future for more complex tasks such as in-air grasping or tool usage, with the support of corresponding end-effectors.}



In this work, we propose a hand-based teleoperation system for omnidirectional drones. Our experiments revealed that using one single mode made it difficult to complete all tasks, \revised{motivating} us to design additional modes. Aerial manipulation tasks can be broadly categorized into two types: long-range operations, which require covering large distances, and precise manipulation, which demands high sensitivity to small input changes. For the former, we develop \textit{Spherical Mode} and \textit{Cartesian Mode}; for the latter, we design \textit{ Operation Mode}. We also propose \textit{Locking Mode}, which can be activated when the operator needs to adjust their visual angle or take a rest. \revised{These modes can be actively chosen by the user through hand gestures to complete a task. During a vertically mounted valve-turning experiment, the operator successfully turned the valve with only one attempt, demonstrating the power of this interaction framework.}

The main contributions of this article are:

\begin{enumerate}
\item We propose a 6-DoF hand-based aerial teleoperation system that can bring the \revised{omnidirectional capability} of human hand into the real world.
\item We design a hand-based interaction framework for this system, including \textit{Spherical Mode} and \textit{Cartesian Mode} for long-range moving, \textit{Operation Mode} for precise operation, \revised{and \textit{Locking Mode} for temporary pauses}.
\item We conducted a \revised{vertically mounted} valve-turning experiment to thoroughly evaluate the proposed interaction framework. \revised{To the best of our knowledge, this is the first time the rotational dexterity of human wrist is intuitively shown on an omnidirectional aerial robot}.
\end{enumerate}

The basic components of hand-based teleoperation system is introduced in Sec. \ref{sec:system}, and the  interaction framework is presented in Sec. \ref{sec:interaction}. We then present the experimental results in Sec. \ref{sec:experiments}, followed by the conclusion in Sec. \ref{sec:conclusion}.

\section{Omnidirectional Aerial Teleoperation}
 \label{sec:system}

\begin{figure}[t]
    \centerline{\includegraphics[trim=5.5cm 2.5cm 9cm 2.5cm,clip,width=3.4in]{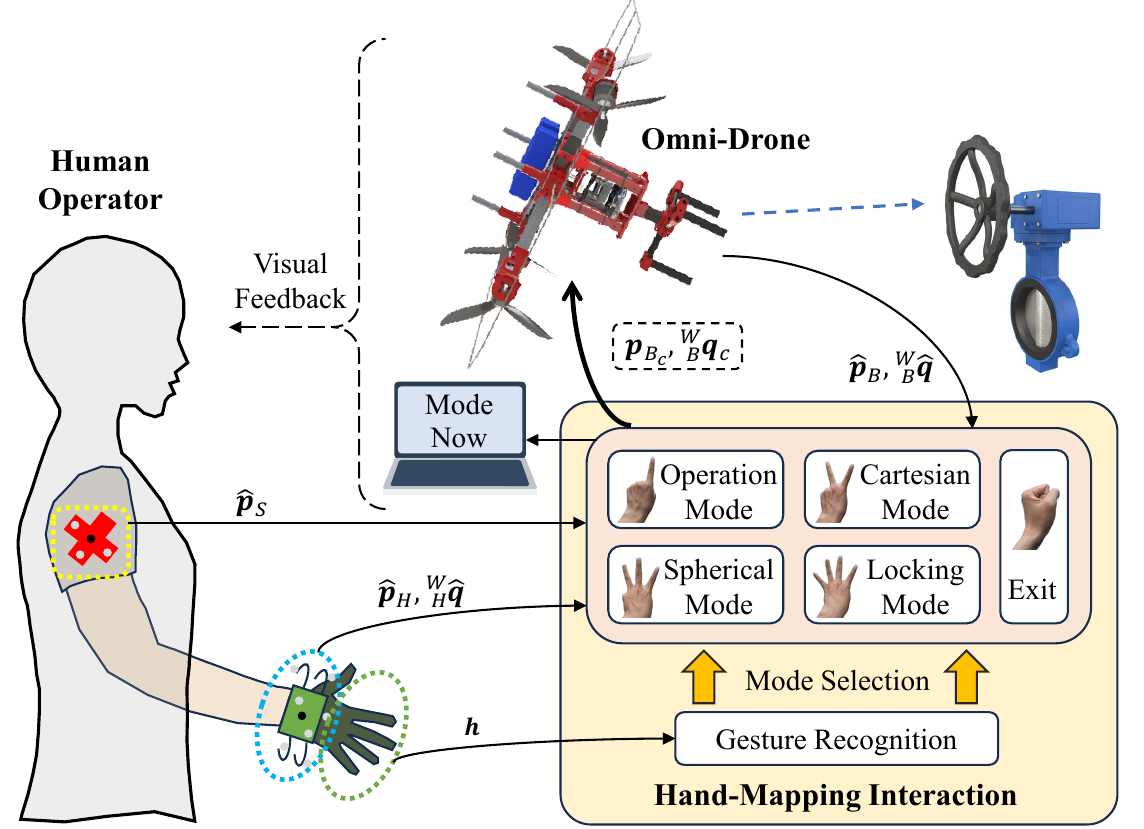}} 
    \vspace*{-1mm}
    \caption{\textbf{Illustration of our teleoperation system.} A human operator wears two marker sets—one on the shoulder and one on the hand—along with a data glove on one hand. The poses of marker sets are tracked for the interaction framework to control the robot, and the glove data are used to recognize hand gestures for mode switching. The teleoperation scenario, including the robot, the operated object, and the current operating mode, is visually fed back to the operator. We report that the feedback of operating mode is crucial for the user to actively arrange all modes for teleoperation.}
    \label{fig:system}
\end{figure}

The teleopration system for an omnidirectional aerial robot consists of five main elements as shown in Fig.\;\ref{fig:system}: (1) \textit{a human operator}, (2) \textit{motion-tracking marker sets} \revised{to localize the shoulder and hand}, (3) \textit{a data glove} \revised{to get hand gestures}, (4) \textit{a screen} to show current control mode, and (5) \textit{an omnidirectional aerial robot}. 
Some of them are introduced in detail in the following sections.

\setlength{\textfloatsep}{8pt plus 1.0pt minus 2.0pt}

\subsection{Notation and Coordinate Systems}

In the following part, we denote scalars in unbold $x, X \in \mathbb{R}$, vectors in bold lowercase $\boldsymbol{x} \in \mathbb{R}^n$, and matrices in bold uppercase $\boldsymbol{X} \in \mathbb{R}^{n \times m}$.
We use $\hat{\cdot}$ to denote estimated values. A vector in $\left\{\mathcal{W}\right\}$ can be denoted as $^{W}\boldsymbol{p}$, and the rotation from $\left\{\mathcal{B}\right\}$ to $\left\{\mathcal{W}\right\}$ is denoted as 
$^{W}_{B}\boldsymbol{q}=\left[q_w, q_x, q_y, q_z\right]^T$ (attitude quaternion). The vector in time $t$ is denoted as $\boldsymbol{p}(t)$.

For coordinate systems, we define the inertial world frame 
\( \left\{\mathcal{W}\right\} = \left\{ ^{W}\boldsymbol{o}, \, ^{W}\boldsymbol{x}, \, ^{W}\boldsymbol{y}, \, ^{W}\boldsymbol{z} \right\} \) 
with its \( ^{W}\boldsymbol{z} \) axis opposite to the direction of gravity. For the sake of simplicity, we omit the symbol for the world frame if one vector is defined under it, i.e., $\boldsymbol{p}$ instead of $^W\boldsymbol{p}$. The coordinate systems for each component are introduced in their subsections.

\subsection{Omnidirectional Aerial Robot}

To thoroughly exploit the omnidirectionality of human hand, an aerial robot that can hover at any orientations is required. Here we use an upgraded version of the tiltable-quadrotor presented in \cite{li_servo_2024}. For the upgrade, we enlarge the rotating limitation of servo modules for omnidirectional flight, and we mount a fork-like end effector on the top of the robot to turn the valve. The control method is actuator-level nonlinear model predictive control, exactly the same with the approach proposed in \cite{li_servo_2024}.

As the same with \cite{li_servo_2024}, we define the body frame $\left\{\mathcal{B}\right\}$ of the drone at its center of gravity (CoG). With the flight controller, the robot can be simplified as a first-order model:
\begin{subequations} \label{eq:rb}
\vspace{-3mm}
\begin{align}
{\dot{\boldsymbol{p}}_B} &= \frac{1}{\boldsymbol{t}_{p}} \left( {\boldsymbol{p}_B}_c - {\boldsymbol{p}_B} \right), \label{eq:rb1} \\[5pt]
{^W_B\dot{\boldsymbol{q}}} &= \frac{1}{\boldsymbol{t}_{q}} \left( {^W_B\boldsymbol{q}^*} \circ {^W_B\boldsymbol{q}_c} \right), \label{eq:rb3}
\end{align}
\end{subequations}
where $\boldsymbol{p}_B$, ${\boldsymbol{p}_B}_c$, ${^W_B\boldsymbol{q}}$, and ${^W_B\boldsymbol{q}}_c$ represent current position, target position, current quaternion, and target quaternion, respectively, $\circ$ refers to quaternion multiplication, $^*$ refers to conjugation, as well as $\boldsymbol{t}_{p}$ and $\boldsymbol{t}_{q}$ denote time constants determined by the performance of controller. The estimated position and orientation are denoted as $\hat{\boldsymbol{p}}_B$ and ${^W_B\hat{\boldsymbol{q}}}$.

The interaction framework \revised{functions as} planning, aiming to calculate ${\boldsymbol{p}_B}_c$ and ${^W_B\boldsymbol{q}}_c$ based on the operator's action.

\subsection{Motion-Tracking Marker Sets}

The motions of operator's shoulder and hand are required in our interaction framework. Specifically, we need the relative direction and distance from the shoulder to the hand to determine the robot's target position. In addition, we require the hand orientation to determine the target orientation. Limited by localization devices, we can \revised{indirectly} obtain these information from the position of the shoulder marker set and the pose of the hand set. The former is defined as $\hat{\boldsymbol{p}}_S$ as well as the latter is defined as $\hat{\boldsymbol{p}}_H$ and ${^W_H\hat{\boldsymbol{q}}}$.


\subsection{Data Glove}

Data gloves can capture the movement of each finger, which is essential for leveraging the dexterity of the human hand. Generally, data gloves are equipped with multiple fabric-based sensors that measure the expansion and contraction of the material as the fingers move. In our system, the stretch values from all knuckles, denoted as $\boldsymbol{h}$, are used to determine the hand gesture.


\section{Hand-Based Interaction Framework}  \label{sec:interaction}

In this section, the proposed control modes for omnidirectional aerial teleoperation are introduced. 
For long-range operations, we introduce \textit{Spherical Mode} and \textit{Cartesian Mode}, which use the spherical and Cartesian coordinate systems, respectively, to interpret the operator's intent. In \textit{Spherical Mode}, the drone behaves as an extension of the operator's arm—moving forward when the hand extends and retracting when the hand is pulled back. In \textit{Cartesian Mode}, the hand functions similarly to a joystick to control the drone's motion. For precise manipulation, we propose \textit{Operation Mode}, which uses a direct mapping strategy based on the position and orientation of operator’s hand. In addition, if the operator needs to adjust their visual perspective during operation, they can activate \textit{Locking Mode} to freeze the drone’s pose. These modes can be jointly used by an operator to finish a task. 
\revised{For mode switching, a data glove is used to capture finger flexion for gesture recognition, with several predefined gestures triggering different modes. Visual feedback of the current control mode is also provided.}

\revised{The teleoperation systems should be resilient to users' incorrect actions, including tremor and involuntary reflexes. To enhance the system robustness, we design stop zones for both \textit{Spherical Mode} and \textit{Cartesian Mode} (Fig. \ref{fig:modes_illustration}) to prevent dangerous motions during long-range movements. Furthermore, we choose to control position instead of velocity to enhance safety. Note that the algorithms can be easily generalized to velocity control for the flight in a wider world.}

\setlength{\textfloatsep}{8pt plus 1.0pt minus 2.0pt}
\begin{figure}[t]
    \centerline{\includegraphics[trim=11.0cm 8.3cm 13.5cm 7cm,clip,width=2.8in]{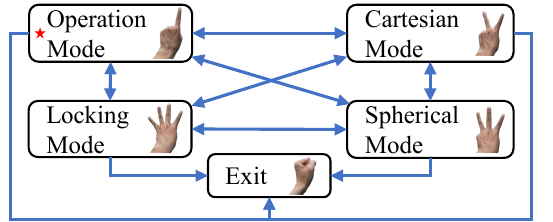}}
    \vspace*{-2mm}
    \caption{\textbf{Mode switching state machine.} The default mode when entering this state machine is \textit{Operation Mode}.}
    \label{fig:mode_switch_logic}
\end{figure}


\subsection{Operation Mode}

In this mode, we aim to establish an intuitive connection between the drone and the operator's hand.
For position, we map the change in hand position to the aerial robot's reference according to the following equation:
\begin{equation}  
\boldsymbol{p}_{B_c} = {\hat{\boldsymbol{p}}_B(t_0)} + \boldsymbol{k} \cdot \left({\hat{\boldsymbol{p}}_H(t)} -  {\hat{\boldsymbol{p}}_H(t_0)} \right),
\end{equation}  
where \( t_0 \) refers to the moment of entering this mode, \(\hat{\boldsymbol{p}}_B(t_0)\) and \(\hat{\boldsymbol{p}}_H(t_0)\) represent the initial positions of the drone and the operator's hand, respectively. $\boldsymbol{k}$ $(\|\boldsymbol{k}\| \le 1)$ \revised{means scale factors for three axes}, where a smaller value indicates a higher precision. Here we set \(\boldsymbol{k}\) to \([1, 1, 1]^T\), and note that the scaling factor \(\boldsymbol{k}\) can be further adjusted according to the required accuracy of the task.
For orientation, the hand attitude is directly mapped to the drone as
\begin{equation}
{{^W_B\boldsymbol{q}}_c} = {^{W}_H\hat{\boldsymbol{q}}.}
\label{eq:rotation_update}
\end{equation}

\subsection{Locking Mode}

When entering this mode at time \( t_0 \), the drone should maintain its pose at that moment:
\begin{equation}  
{\boldsymbol{p}_B}_c = {\hat{\boldsymbol{p}}_B(t_0)},
\end{equation} 
\begin{equation}
{{^W_B\boldsymbol{q}}_c} = {^{W}_B\hat{\boldsymbol{q}}(t_0).}
\end{equation}
This allows the operator to freely adjust their viewing perspective or perform other tasks without affecting the drone, \revised{providing a start point for further operations}.

\subsection{Moving Modes}

\setlength{\dbltextfloatsep}{8pt plus 1.0pt minus 2.0pt}
\begin{figure}[tb] 
    \centering
    \subfloat[Design of \textit{Spherical Mode}. Specifically, the robot's movement is defined in a spherical coordinate system, with the shoulder as the origin and the line from the shoulder to the hand as the polar axis. The robot remains on this axis, and its distance from the origin is determined by the \revised{hand-to-shoulder distance}.]{
        \includegraphics[trim=0cm 1.5cm 0cm 0cm,clip,width=3.4in]{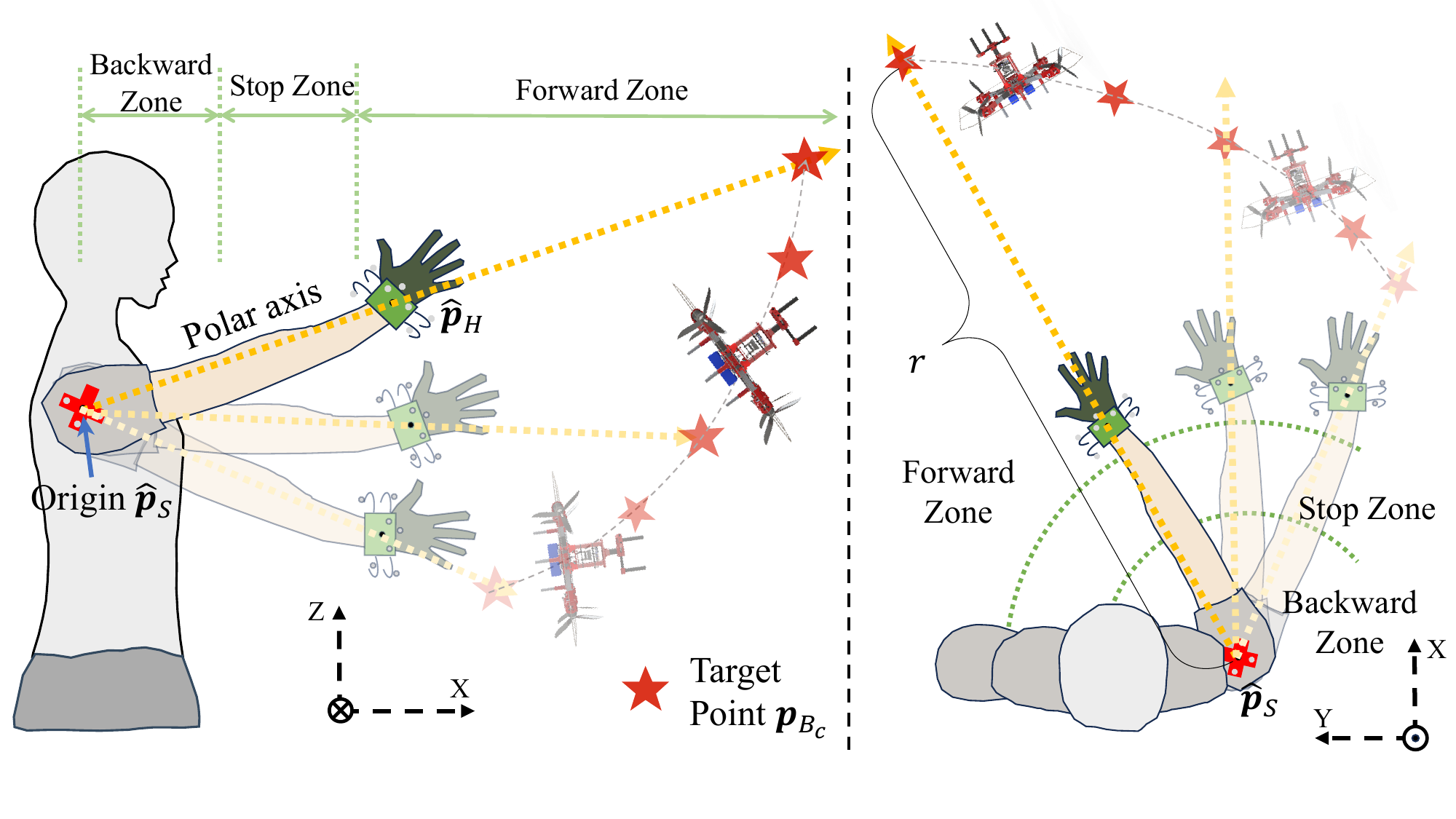}
        \label{fig:spherical_mode_illustration}
    }\\ \vspace{-1mm}
    \subfloat[Design of \textit{Cartesian Mode}. Specifically, a point in front of the shoulder serves as the origin. The robot remains stationary when near this point and inside the \textit{Stop Zone}. When outside the zone, the robot's fly direction aligns with the line from the origin to the hand, resembling joystick-based control.]{
        \includegraphics[trim=2.5cm 1.5cm 2cm 2.5cm,clip,width=3.3in]{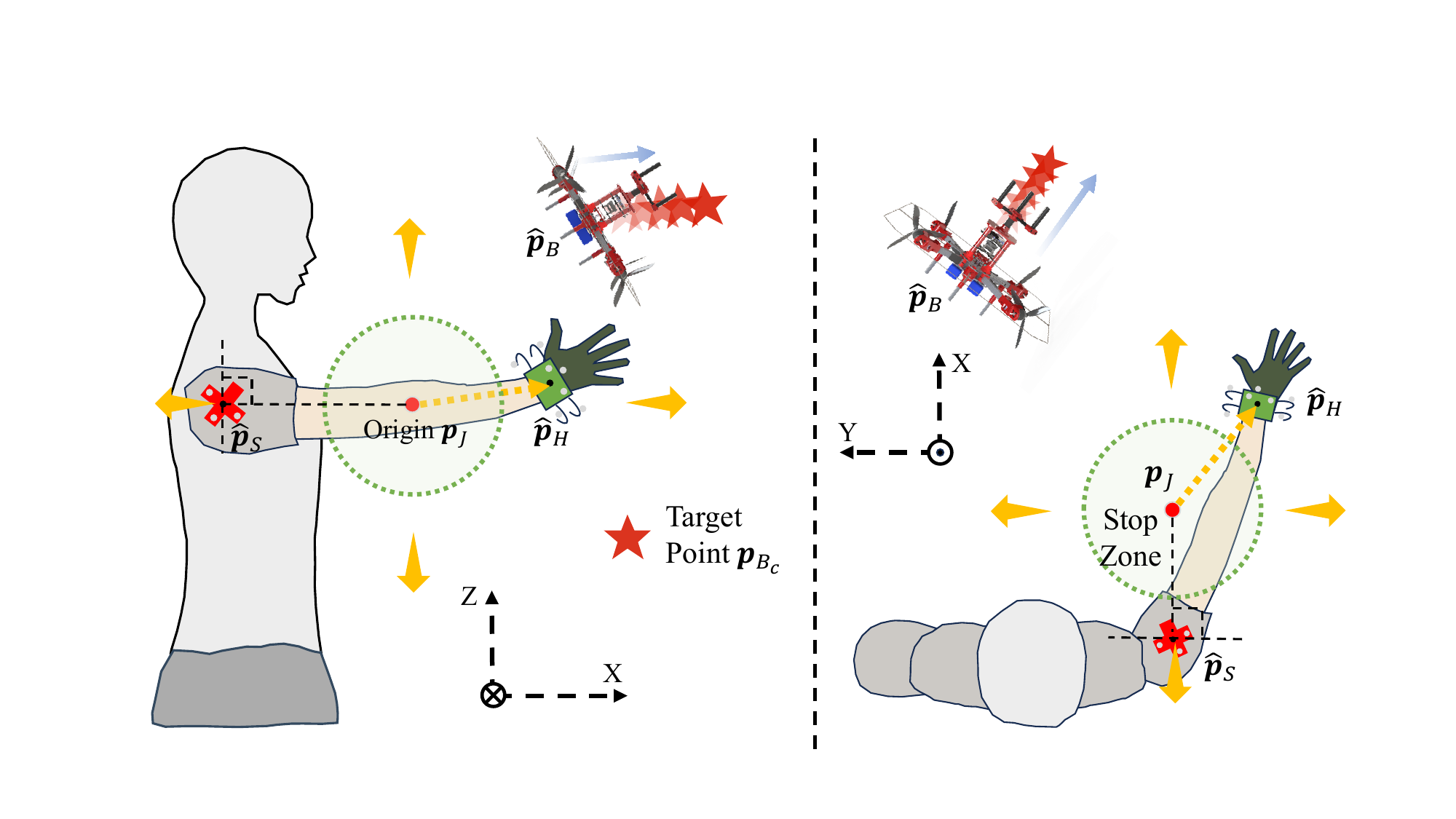}
        \label{fig:cartesian_mode_illustration}
    }
    \caption{\revised{\textbf{Illustrations of \textit{Spherical Mode} and \textit{Cartesian Mode}}}}
    \vspace{-0.1cm}
    \label{fig:modes_illustration}
\end{figure}

\subsubsection{Spherical Mode}

The geometric explanation of \textit{Spherical Mode} is illustrated in Fig.\;\ref{fig:spherical_mode_illustration}. 
In this mode, the shoulder is considered as the origin of the spherical coordinate system, with the vector from the shoulder to the hand defining the polar axis. A target point \({\boldsymbol{p}_{B_c}}\) lies along the polar axis, and the drone always moves towards this position. The radial distance \(r\in \mathbb{R}\) from the origin to the target point is controlled by \revised{the distance between operator's hand and shoulder}. When the hand and shoulder are within a certain range, the radial distance \revised{from the origin to robot} remains unchanged, \revised{leaving the operator a chance to rest}. As the hand moves away from the shoulder, the radial distance increases; when the hand moves closer, the radial distance decreases. This process is detailed in Algorithm\;\ref{algo:1}, which is continuously executed to update the target position. The target orientation remains the direct mapping as (\ref{eq:rotation_update}).

\begin{algorithm}[bt]
\caption{Drone Position Update in \textit{Spherical Mode}}
\label{algo:1}
\label{alg:polar_mode}
\begin{algorithmic}[1]

\Require 1) \( \hat{\boldsymbol{p}}_H \), the hand position; 2) \( \hat{\boldsymbol{p}}_S \), the shoulder position; 3) \( r \), the radial distance between the target point and the shoulder; 4) \( r_{\text{max}}, r_{\text{min}} \), the maximum and minimum values for \( r \); 5) \( d_{\text{min}}, d_{\text{max}} \), the minimum and maximum distance thresholds for adjusting \( r \); 6) \( \Delta r \), the adjustment increment for \( r \).
\Ensure \(\boldsymbol{p}_{B_c}\), the target position of the aerial robot.
\While{\(\text{in } \textit{Spherical Mode}\)}  \Comment{\revised{run at 50\;\si{Hz}}}
\State \( \boldsymbol{d}_{s \rightarrow h} = {\hat{\boldsymbol{p}}_H} - {\hat{\boldsymbol{p}}_S} \); 
\State \( {\vec{\boldsymbol{d}}} = \frac{\boldsymbol{d}_{s \rightarrow h}}{{\| \boldsymbol{d}_{s \rightarrow h}\|}} \); 
\If{\(\| \boldsymbol{d}_{s \rightarrow h}\|\) < \( d_{\text{min}} \)}
    \State \( r \leftarrow r - \Delta r \); 
\ElsIf{\(\| \boldsymbol{d}_{s \rightarrow h}\|\) > \( d_{\text{max}} \)}
    \State \( r \leftarrow r + \Delta r \); 
\EndIf
\State \( r \leftarrow \max\left( \min\left( r, r_{\text{max}} \right), r_{\text{min}} \right) \); 
\State \( {\boldsymbol{p}_{B_c}} \leftarrow {\hat{\boldsymbol{p}}_S} + r \cdot {\vec{\boldsymbol{d}}} \); 
\EndWhile
\end{algorithmic}
\end{algorithm}


\subsubsection{Cartesian Mode}

The geometric explanation of \textit{Cartesian Mode} is depicted in Fig.~\ref{fig:cartesian_mode_illustration}. Once entering this mode, a new local coordinate system \( \left\{\mathcal{J}\right\} \) is generated, with its origin \( {\boldsymbol{p}_J} \) located 0.3\;\si{m} along the world X-axis from \( {\hat{\boldsymbol{p}}_S} \). The directions of X, Y, and Z axes are aligned with those of the world coordinate system. Starting from the origin \( \boldsymbol{p}_J \), the vector \( \boldsymbol{d} \) points to \( \hat{\boldsymbol{p}}_H \), and its unit vector \(\vec{\boldsymbol{d}}\) provides a normalized reference for drone motion. When the distance between \( {\hat{\boldsymbol{p}}_H} \) and \( {\hat{\boldsymbol{p}}_J} \) exceeds a threshold, the drone updates its target position and moves a fixed distance in a direction parallel to \( \vec{\boldsymbol{d}} \). This process is detailed in Algorithm\;\ref{alg:cartesian_mode}. The target orientation is updated in the same way as the \textit{Operation Mode} via (\ref{eq:rotation_update}).

\begin{algorithm}[bt]
\caption{Drone Position Update in \textit{Cartesian Mode}}
\label{algo:2}
\label{alg:cartesian_mode}
\begin{algorithmic}[1]
\Require 1) \( \hat{\boldsymbol{p}}_H \), the hand position; 
          2) \( \hat{\boldsymbol{p}}_B \), the robot position; 
          3) \( \hat{\boldsymbol{p}}_S \), the shoulder position; 
          4) \( d_{\text{thresh}} \), the threshold for updating the drone's target position; 
          5) \( \Delta d \), the fixed distance the drone moves in one round.
\Ensure \( \boldsymbol{p}_{B_c} \), the target position of the aerial robot.
\While{\(\text{in } \textit{Cartesian Mode}\)}   \Comment{\revised{run at 50\;\si{Hz}}}
\State \( \boldsymbol{p}_J = \boldsymbol{\hat{p}}_{S} + [0.3, 0, 0]^T\);
\State \( \boldsymbol{d}_{j \rightarrow h} = {\hat{\boldsymbol{p}}_H} - {\boldsymbol{p}}_J\);
\State \( \vec{\boldsymbol{d}} = \frac{\boldsymbol{d}_{j \rightarrow h}}{\| \boldsymbol{d}_{j \rightarrow h} \|} \);
\If{\( \| \boldsymbol{d}_{j \rightarrow h} \| > d_{\text{thresh}} \)}
    \State \( \boldsymbol{p}_{B_c} \leftarrow {\hat{\boldsymbol{p}}_B} + \Delta d \cdot \vec{\boldsymbol{d}} \);
\Else
    \State \( \boldsymbol{p}_{B_c} \leftarrow {\hat{\boldsymbol{p}}_B}\);
\EndIf
\EndWhile
\end{algorithmic}
\end{algorithm}

\subsection{Mode Switch}

We use a data glove to switch control modes by mapping specific hand gestures to their corresponding modes.
The glove measures finger flexion, and higher value indicates more contraction. We evaluate the contraction of each finger: if its flexion exceeds a predetermined contraction threshold, it is considered contracted; if it falls below an extension threshold, it is considered extended; and the values in between are considered undetermined. \revised{When the contraction states of all five fingers match a predefined pattern shown in Fig.\;\ref{fig:mode_switch_logic}}, the gesture is recognized. If the gesture is maintained for a period of time, the control mode is accordingly switched.

\setlength{\textfloatsep}{8pt plus 1.0pt minus 2.0pt}
\begin{figure}[t]
    \centerline{\includegraphics[trim=7.0cm 6cm 11.5cm 6.5cm,clip,width=3.0in]{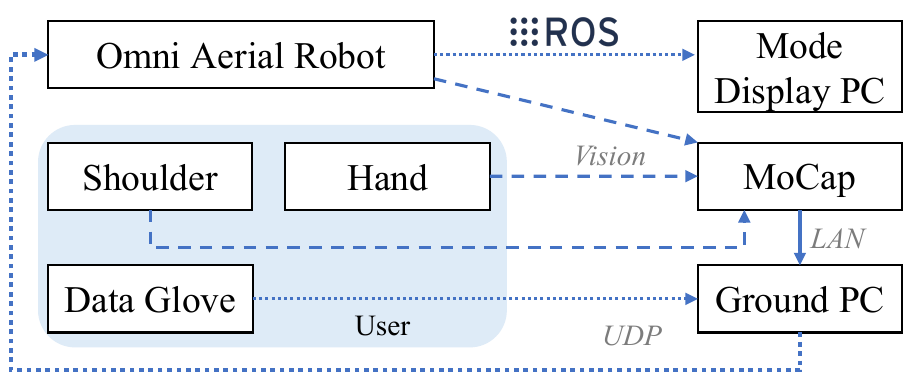}}
    \vspace*{-3mm}
    \caption{\textbf{Communication workflow of our teleoperation system.}}
    \vspace{-1mm}
    \label{fig:hardware_connections}
\end{figure}

\subsection{Feedback for Human}


During our preliminary experiments, we observe that operators \revised{struggle to perceive control mode transitions}.
This lack of situational awareness significantly reduces the perceived safety of the human-robot interaction. To address this issue, we implement a visual feedback mechanism by positioning a display within the operator’s line of sight. The display presents the current control mode as text accompanied by a unique background color, thereby enhancing the operator’s awareness of the control mode. \revised{Simultaneously, the robot should always be within the operator’s field of vision for flight safety.}



\section{Experiments}  \label{sec:experiments}

\subsection{Setup}



\subsubsection{Platform}

We developed a teleoperation system for an omnidirectional aerial robot with the main components shown in Fig.\;\ref{fig:teleop_real}. 
The robot weighs 3.08\;\si{kg} and features a wheelbase of 0.55\;\si{m}. It is controlled by a Khadas VIM4 computer running Ubuntu 20.04, where all algorithms are executed. In comparison with the hardware described in \cite{li_servo_2024}, we upgrade the servo from the Kondo KRS-3302 to the Dynamixel XC330-T181-T, enabling rotation beyond 180 degrees. \revised{Regarding the state estimation, since the core of this research is the concept of intuitive aerial teleoperation, we use an OptiTrack Motion Capture system to obtain the poses. The motion marker sets for the shoulder and hand are custom-designed and 3D printed, where reflective balls are affixed for localization.} Finally, a Studio Glove from StretchSense is used to track hand gestures.


The communication architecture of our teleoperation system is illustrated in Fig.\;\ref{fig:hardware_connections}. The reflective balls on the aerial robot, along with the shoulder and hand marker sets, are captured by the motion capture cameras using reflected infrared light. These data are then transmitted to a ground computer, where the Motive software computes the poses of the tracked objects. Simultaneously, data from the glove is sent to the ground PC, where Hand Engine Lite calculates the finger gestures following calibration. The resulting pose and finger-bending information are wirelessly transmitted to the aerial robot via UDP.
During flight, all the algorithms are executed onboard and the library SMACH is used for state machine. The current control mode is sent to a secondary laptop for visualization.

\revised{Regarding parameters, for \textit{Spherical Mode} we choose $r_{\rm min}=1 $\;\si{m}, $r_{\rm max}=20 $\;\si{m}, $d_{\rm min}=0.2 $\;\si{m}, $d_{\rm max}=0.4 $\;\si{m}, and $\Delta r = 0.01$\;\si{m}; for \textit{Cartesian Mode} we choose $d_{\rm thresh} = 0.15$\;\si{m} and $\Delta d = 0.01$\;\si{m}. All loops are run at 50\;\si{Hz}.}


\subsubsection{Procedures}

\begin{figure}[t] 
    \centering
    \subfloat[The operator is using \textit{Spherical Mode} to bypass an obstacle.]{
        \includegraphics[trim=5.5cm 4.0cm 1cm 3.5cm,clip,width=3.3in]{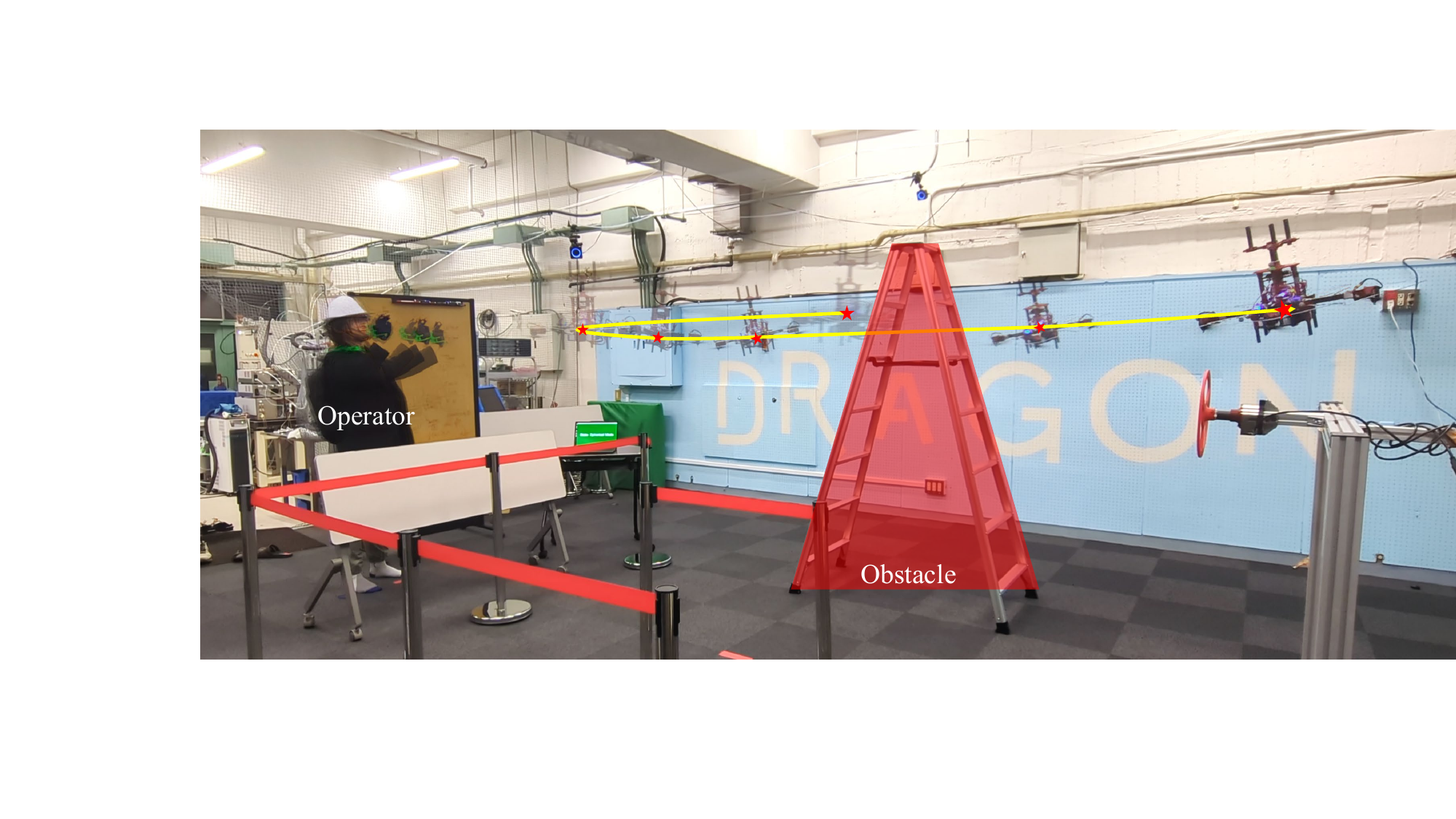}
        \label{fig:spherical_mode_exp}
    }\\ 
    \subfloat[\revised{Hand trajectory.}]{
    \includegraphics[trim=1mm 0 0 0,clip,width=1.5in]{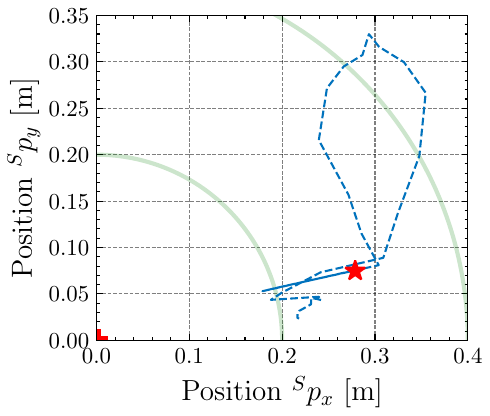}
    \label{fig:spherical_hand_data}
    }
    \subfloat[\revised{Flight trajectory.}]{
    \includegraphics[trim=0 0 0 0,clip,width=1.68in]{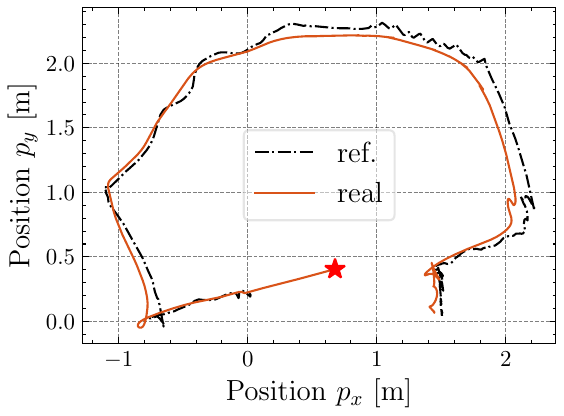}
    \label{fig:spherical_traj_data}
    }
    \caption{\textbf{Results for \textit{Spherical Model}.} The red star denotes the start point, the red cross denotes the shoulder, and the green lines denotes the boundary of backward, stop, and forward zones. \revised{See Fig. \ref{fig:spherical_mode_illustration} for details.The horizontal tracking RMSEs are X: 0.2248\;\si{m} and Y: 0.1564\;\si{m}.}}
    \vspace{-0.1cm}
    \label{fig:spherical_exp_result}
\end{figure}

\begin{figure}[t] 
    \centering
    \subfloat[The operator is using \textit{Cartesian Mode} to fly through a right-angle turn.]{
        \includegraphics[trim=10.0cm 5.0cm 0 3.0cm,clip,width=3.3in]{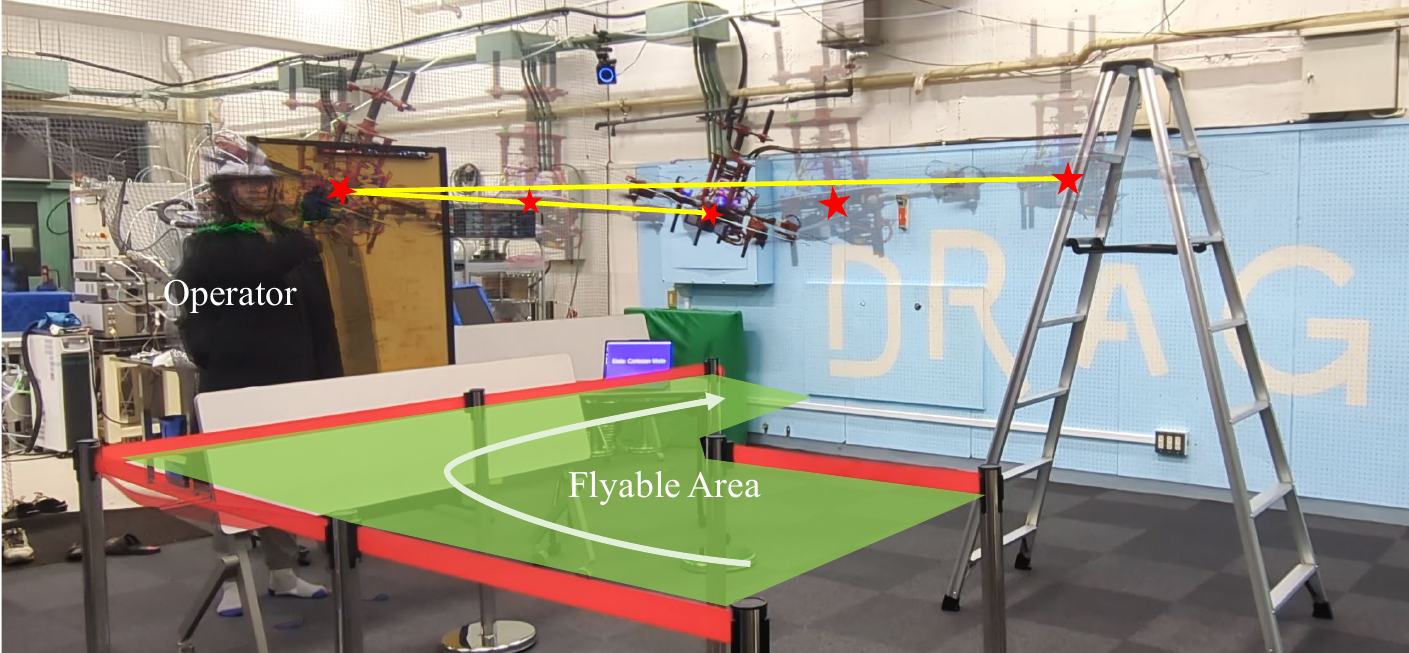}
        \label{fig:cartesian_mode_exp}
    }\\
    \subfloat[\revised{Hand trajectory.}]{
    \includegraphics[trim=1mm 0 0 0,clip,width=1.15in]{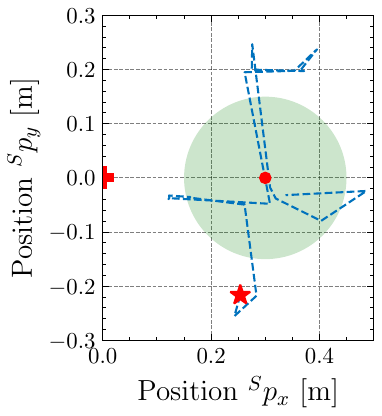}
    \label{fig:cartisesian_hand_data}
    }
    \subfloat[\revised{Flight trajectory.}]{
    \includegraphics[trim=0 0 0 0,clip,width=1.9in]{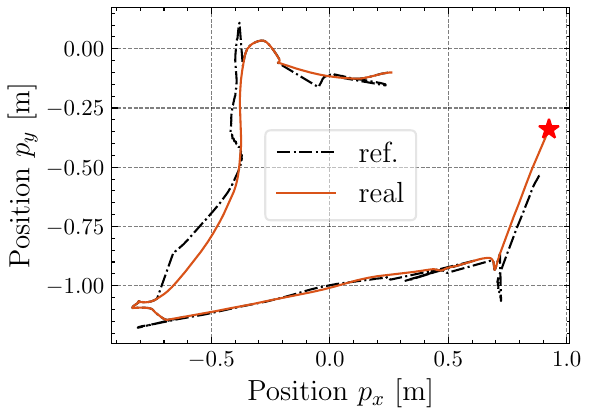}
    \label{fig:cartesian_traj_data}
    }
    \caption{\textbf{Results for \textit{Cartesian Mode}.} The red star denotes the start point, the red cross denotes the shoulder, and the green area means the stop zone. \revised{The horizontal tracking RMSEs are X: 0.1009\;\si{m} and Y: 0.0962\;\si{m}.}}
    \vspace{-0.1cm}
    \label{fig:cartesian_exp_result}
\end{figure}

The experiment is conducted in a field \revised{of} 7\;\si{m} in length, 6\;\si{m} in width, and 2\;\si{m} in height. To enhance safety, the height control is simplified by directly mapping the hand's height to the robot's target altitude. To thoroughly evaluate the interaction framework, we design a mission that involves obstacle avoidance, valve turning, and corridor navigation.
Specifically, the operator must guide the drone to a valve area while avoiding a ladder (Fig.\;\ref{fig:spherical_mode_exp}). Next, the operator overcomes occlusion challenges to align the fork effector with the vertically installed valve, turn it, and then return to the starting position (Fig.\;\ref{fig:mapping_locking_exp}). Finally, the operator directs the drone out of the task area along a path that includes a right turn (Fig.\;\ref{fig:cartesian_mode_exp}). Throughout this process, the current control mode is displayed on a laptop for feedback. 
The operator is required to select the appropriate control mode at the right time to successfully complete the task.


\subsection{Results}

Although the operator changed the control mode through \textit{Spherical Mode} → \textit{Operation/Locking Mode} → \textit{Cartesian Mode}, we first present the results for long-range movement, followed by those for precise control.

\subsubsection{Long-Range Moving}

The operator employs \textit{Spherical Mode} to enter the task area and switches to \textit{Cartesian Mode} to leave. The corresponding scenarios are illustrated in Fig.\;\ref{fig:spherical_mode_exp} and Fig.\;\ref{fig:cartesian_mode_exp}, respectively. \textit{Spherical Mode} is intuitive, as it feels like the extension of arm movements, while \textit{Cartesian Mode} is well-suited for navigating straight-line paths in real-world environments.

The hand trajectories for each mode are shown in Fig.\;\ref{fig:spherical_hand_data} and Fig.\;\ref{fig:cartisesian_hand_data}, where the ranges of hand movement in two modes are within 0.45\si{m}, with \textit{Cartesian Mode} being slightly smaller than \textit{Spherical Mode}. The flight trajectories for each mode are depicted in Fig.\;\ref{fig:spherical_traj_data} and Fig.\;\ref{fig:cartesian_traj_data}. In \textit{Spherical Mode}, the operator successfully guides the drone around obstacles to reach the target. In \textit{Cartesian Mode}, although the trajectory does not perfectly align with the centerline of the turn, the robot consistently remains within the corridor.

\subsubsection{Accurate Operation}

\begin{figure}[t]
    \centerline{\includegraphics[trim=1.5cm 3.8cm 1.5cm 0,clip,width=3.4in]{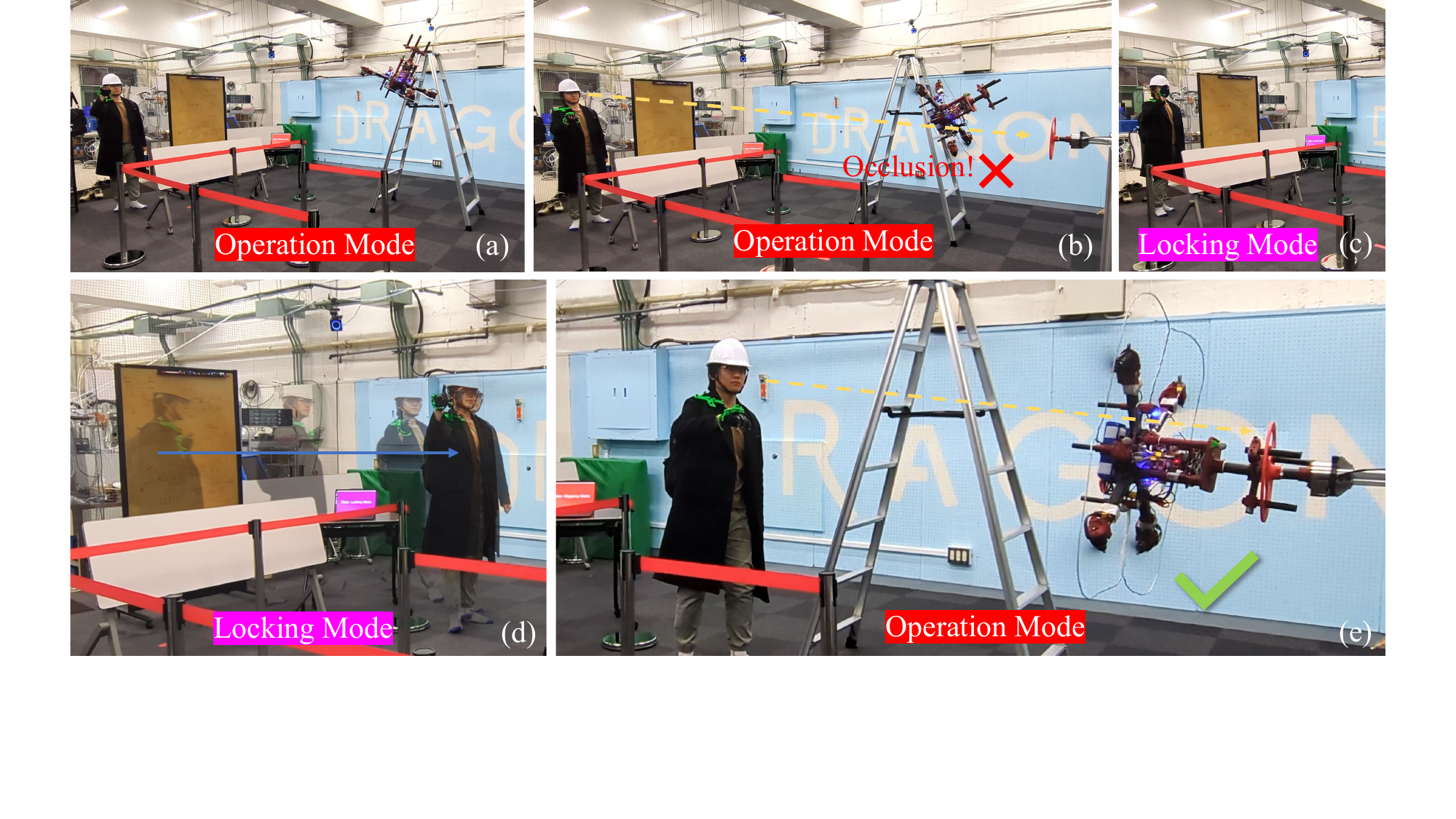}}
    \vspace*{-2mm}
    \caption{\textbf{Scenarios for valve-turning.} The operator switches to \textit{Operation Mode} for precise operation, but their line of sight is occluded by the robot. To resolve this, they switches to \textit{Locking Mode}, moves to a different position, switches back to \textit{Operation Mode}, and succeeds.}
    \label{fig:mapping_locking_exp}
\end{figure}

\begin{figure}[t]
    \centerline{\includegraphics[trim=0 3mm 0 0,clip,width=3.0in]{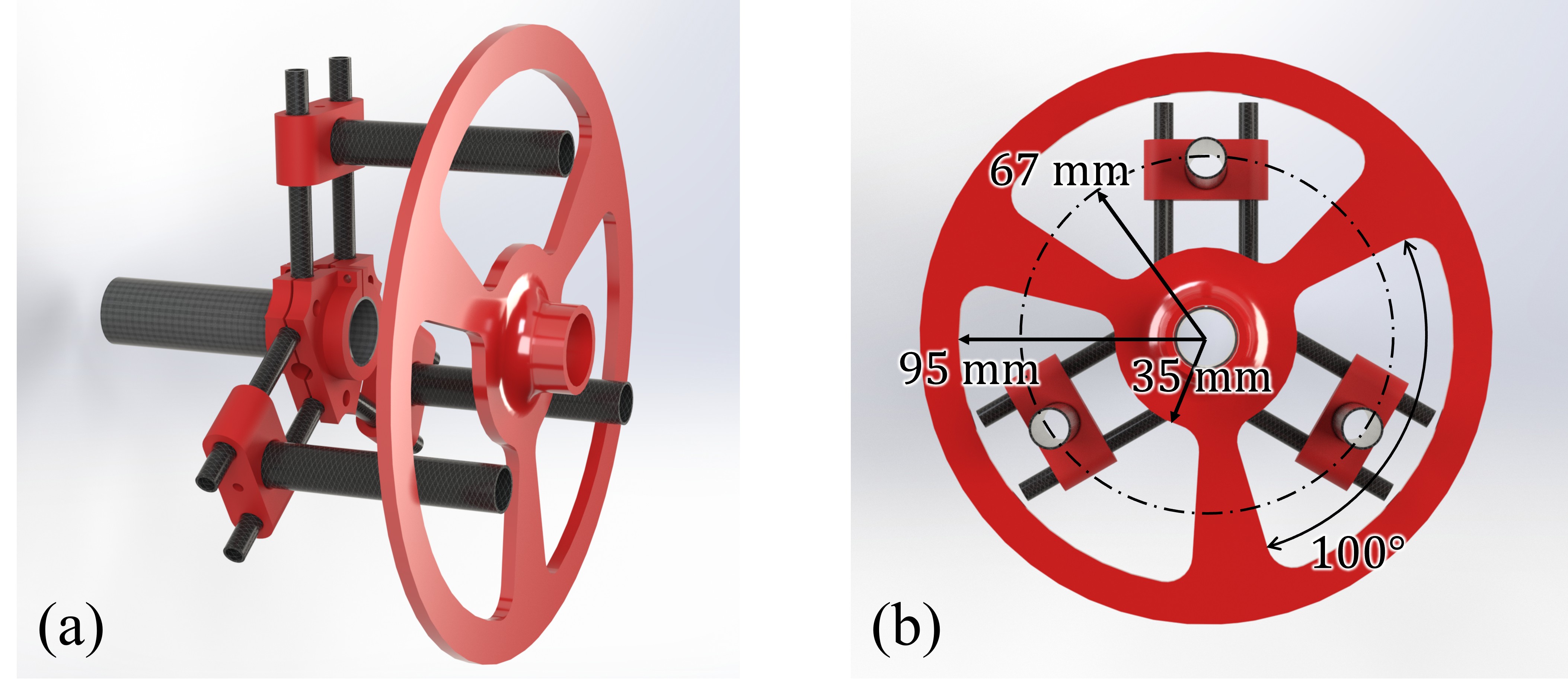}}
    \vspace*{-2mm}
    \caption{\revised{\textbf{Size illustration of a valve and a fork-like end-Effector.} The valve size is designed from a real product, and the radius of end-effector is around half of the feasible radius to ease docking. The sizes leave positional and rotational redundancy as around 30\;\si{mm} and 90$^\circ$, respectively.}}
    \label{fig:valve_size}
\end{figure}

\begin{figure}[t]
    \centerline{\includegraphics[trim=0 1mm 0 1mm,clip,width=3.4in]{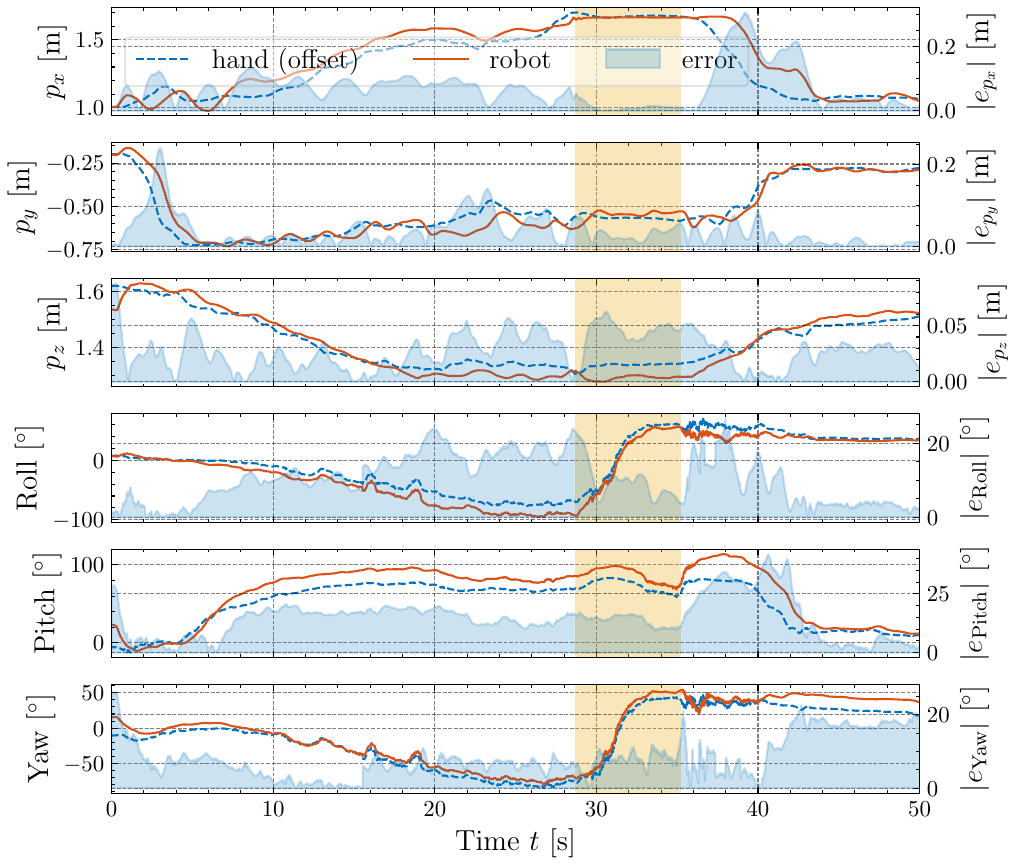}}
    \vspace*{-3mm}
    \caption{\revised{\textbf{Results for the accurate valve-turning task.} 
    The hand poses with initial offset is set as reference, and the teleoperation latency is approximately 0.3\textasciitilde0.5\;\si{s}.
    The yellow region highlights the valve-turning process. The absolute error increases right after the yellow area when the end effector is pulled away from the valve. \revised{The RMSEs are $p_x$: 0.0930\;\si{m}, $p_y$: 0.0519\;\si{m}, $p_z$: 0.0321\;\si{m}; $q_w$: 0.0944, $q_x$: 0.0467, $q_y$: 0.1122, $q_z$: 0.0816; roll: 7.1949$^\circ$, pitch: 11.6848$^\circ$, yaw: 5.9168$^\circ$, where the Euler angle error is converted from quaternion error for readability.}}}
    \label{fig:mapping_mode_traj}
\end{figure}

As depicted in Fig.\;\ref{fig:mapping_locking_exp}, once the drone enters the task area, the operator switches to \textit{Operation Mode} for precise control. Given the extended working range of aerial manipulation, the distance between the robot and the object is often much shorter than that between the robot and the operator, leading to occlusion issues. To overcome this, the operator switches to \textit{Locking Mode} and moves around to adjust their perspective. This mode transition ultimately enables the operator to successfully turn the valve. \revised{The sizes of the valve and end-effector are illustrated in Fig.\;\ref{fig:valve_size}, indicating the requirement of operation accuracy in this task.
The trajectories of both the operator's hand and the robot as well as the tracking errors are shown in Fig.\;\ref{fig:mapping_mode_traj}.}

\revised{The successful completion of the valve turning task reflects the necessity of human in the loop. Actually, the robot's controller has imperfections as shown in Fig.\;\ref{fig:mapping_mode_traj}: the RMSEs of all positional axes are over 30\;\si{mm}, which are larger than the maximum tolerance (Fig.\;\ref{fig:valve_size}) to insert the end-effector into the valve. Considering the flight seems stable enough in Fig.\;\ref{fig:mapping_mode_traj}, these large RMSEs majorly come from steady error. Simultaneously, a steady attitude error is observed, indicating model inaccuracies in the NMPC controller.
Furthermore, the orientation of the human hand is mapped to the robot's CoG frame instead of the end-effector frame, increasing the difficulty of insertion.
Eventually, the teleoperation system exhibits a latency of about 0.3\textasciitilde0.5\;\si{s}.
However, the operator could still overcome all these drawbacks and finish the task within one attempt, showcasing the robustness of the proposed teleoperation system. Actually, if only the flight is stable, the operator cannot feel the steady error in control, as they directly do visual servoing from the robot to the valve. On the contrary, solving the coordinate transformation problem should reduce the operating difficulty, reported by the operator. Finally, the control latency is adequate for this low-speed valve-turning task, but it can be improved by predictive control and should be meaningful to a wider variety of tasks.
}


\subsection{Discussions}

In \textit{Spherical Mode}, operators found the control experience similar to moving their own hand; however, accurately knowing the radial distance to the target and precisely navigating around obstacles remained challenging. In contrast, \textit{Cartesian Mode} enhanced operators' confidence in understanding the drone’s fly direction, though it felt less natural than the spherical approach.
\revised{Controlling orientation in these modes was reported not particularly helpful for the valve-turning task, so the robot could remain horizontal flight during long-range moving to save energy.}
\textit{Operation Mode} provided the most natural and intuitive control experience overall, despite the lack of tactile feedback and other sensory inputs, which prevented users from knowing when the robot touched an object. The reduced spatial awareness in the operating environment further hampered precise operation, underscoring the need of \textit{Locking Mode} to adjust perspective—a function that could also be achieved by mounting a camera on the robot. Lastly, \revised{feeding back current control mode via a screen} significantly aided operators in actively selecting the most appropriate mode during operation.

\setlength{\textfloatsep}{8pt plus 1.0pt minus 2.0pt}

\setlength{\textfloatsep}{8pt plus 1.0pt minus 2.0pt}

\setlength{\textfloatsep}{8pt plus 1.0pt minus 2.0pt}


\section{Conclusion}  \label{sec:conclusion}

In this work, we proposed an \revised{intuitive} aerial teleoperation system for omnidirectional aerial robots that brings the flexibility of human hands into aerial environments. Based on this system, we designed an interaction framework including \textit{Spherical Mode}, \textit{Cartesian Mode}, \textit{Operation Mode}, and \textit{Locking Mode}, \revised{which were jointly evaluated in a vertically mounted valve-turning experiment.}
We hope the interaction framework defined in this work can serve as a valuable reference for the aerial manipulation community.

\revised{Future work first includes fully onboard state estimation (e.g., using GPS or LiDAR) for both the robot and wearable devices, aiming to enable teleoperation in the wild.
In addition, a gimbal-mounted camera can be used to stream first-person video for remote teleoperation beyond visual range \cite{hatano_improved_2025}.
Another challenge in remote teleoperation is signal delay, which could be mitigated through human behavior prediction.
Moreover, visual feedback from the robot could be upgraded to force feedback, enabling bilateral teleoperation. The data gloves can also be enhanced with haptic devices \cite{zhu_haptic-feedback_2020} to better capture human intent.
Finally, supported by various end-effectors, hand gestures can be extended to more complex tasks (such as in-air grasping or tool usage), which may require alternative interaction methods, including \textit{shared control} \cite{selvaggio_autonomy_2021} between human and machine. This is particularly important for compound rotations that exceed the natural flexibility of the human wrist and forearm.}










\bibliographystyle{bibtex/IEEEtran}
\bibliography{bibtex/IEEEabrv, bibtex/my_config, bibtex/references_pure}


\end{document}